# Sketch Recognition using Domain Classification

Vasudha Vashisht
Assistant Professor
Dept of Computer Sc & Engg
Lingaya's University
Faridabad, Haryana, INDIA

Tanupriya Choudhury
Senior Lecturer
Dept of Computer Sc & Engg
Lingaya's University
Faridabad, Haryana, INDIA

Dr. T. V. Prasad
Dean (R&D)
Lingaya's University
Faridabad, Haryana, INDIA

*Abstract*— Conceptualizing away the sketch processing details in a user interface will enable general users and domain experts to create more complex sketches. There are many domains for which sketch recognition systems are being developed. But, they entail image-processing skill if they are to handle the details of each domain, and also they are lengthy to build. The implemented system's goal is to enable user interface designers and domain experts who may not have proficiency in sketch recognition to be able to construct these sketch systems. This sketch recognition system takes in rough sketches from user drawn with the help of mouse as its input. It then recognizes the sketch using segmentation and domain classification; the properties of the user drawn sketch and segments are searched heuristically in the domains and each figures of each domain, and finally it shows its domain, the figure name and properties. It also draws the sketch smoothly. The work is resulted through extensive research and study of many existing image processing and pattern matching algorithms.

## I. INTRODUCTION

As computers become an integral part of our lives, it becomes increasingly important to make working with them easier and more natural. It is visional to make human-computer interaction as easy and as natural as human-human interaction. As part of this vision, it is imperative that computers understand forms of human-human interaction, such as sketching. Computers should be able to understand the information encoded in diagrams drawn by and for scientists and engineers. A mechanical engineer, for example, can use a hand-sketched diagram to depict his design to another engineer. Sketching is a natural modality of human-computer interaction for a variety of tasks [7].

In an attempt to combine the freedom provided by a paper sketch with the powerful editing and processing capabilities of an interpreted diagram, sketch recognition systems have been developed for many domains, including Java GUI creation, UML class diagrams, and mechanical engineering. Sketch interfaces:

1) Interact more naturally than traditional mouse-and-palette tools by allowing users to hand-sketch diagrams,
2) Can connect to a back-end system (such as a CAD tool) to offer real-time design advice,
3) Recognize the shape as a whole to allow for more powerful editing,
4) Beautify diagrams, removing mess and clutter, and thereby
5) Notify the sketcher that the shapes have been recognized correctly[6].

Previous sketch systems required users to learn a particular stylized way of drawing, and used a feature-based recognition algorithm, such as a Rubine or a GraffitiTM-type algorithm. What these algorithms lose in natural interaction by requiring the sketcher to draw in a particular style, they gain in speed and accuracy. Rather than recognizing shapes, the algorithm recognizes sketched gestures, where each gesture represents a single shape. These sketched gestures focus more on how something was drawn than on how the drawn object looks. These recognition algorithms require that each gesture can be drawn in a single stroke in the same manner (i.e., same underlying stylistic features–stroke direction, speed, etc.) each time. Each gesture has recognized based on a number of features of that stroke, such as the initial angle of the stroke, end angle, speed, number of crosses, etc. Because of these requirements, the gesture representing the shape may look different from the shape itself.

Further it is noted that :

- Human generated descriptions contained syntactic and conceptual errors, and that
- It is more natural for a user to specify a shape by drawing it than by editing text[7].

When working in a closed domain such as this one, the computer knows exactly which conceptual uncertainties remain, and which hypotheses need to be tested and confirmed. The system builds a shape description language, using a modification of the version spaces algorithm that handles interrelated constraints.

To achieve the goals, the system is implemented for the segmentation of the sketch. This system takes the input from the user about the positions where the sketch has drawn and then it process this information to divide the sketch into a number of segments according to the position and the direction of the sketch that has drawn by the user.

To allow for natural drawing in proposed sketch recognition systems, shapes are described and recognized in terms of the subshapes that make up the shape and the geometric relationships (constraints) between the subshapes. Strokes are first broken down into a collection of primitive shapes using a SEGMENTATION technique which operates on the pixels and their orientation. A higher-level shape is then recognized by searching for possible subshapes in the domain and testing that the appropriate geometric constraints hold. The geometric constraints confirm orientation, angles, relative size, and relative location.

The remainder of this paper is organized as follows. Section 2 briefly reviews the basics of The User Interface Development Framework with Domain specific information. Section 3 gives the details of methodology by specifically talking about the language for describing drawing and display, its building blocks and limitations. Section 4 of the paper discusses the overall recognition system, the algorithm used along with its limitations and further research directions.

## II. USER INTERFACE DEVELOPMENT

### A. Domain-specific Information

When constructing a user interface, the domain-specific information is able to be obtained by asking the following questions [7]:

- What are the observable states to be recognized?
- How are these states to be recognized?
- What should happen when these states are recognized?
- How can we modify these states?

In sketch recognition user interfaces, the domain-specific information is obtained by asking these questions:

- What shapes are in the domain?
- How is each shape recognized?
- What should happen after each shape is recognized?[8]

Many domain-specific events can occur after a shape is recognized, but what is common in most domains is a change in display. Sketchers often prefer to have a change in display to confirm that their object was recognized correctly, as a form of positive feedback. Changes in display may also function as a way to remove clutter from the diagram. For example, the system may replace several messy hand-drawn strokes with a small representative image[4]. A change in the display may vary from a simple change in color, a moderate change of cleaning up the drawn strokes (e.g. straightening lines, joining edges), to a more drastic change of replacing the strokes with an entirely different image. Because display changes are so popular and so common to most domains, so they are included in the language.

This framework not only defines which shapes are in the domain and how they are to be recognized in the domain, it also recognizes the importance display in creating an effective user interface. Developers of different domains may want the same shape to be displayed differently: Compare a brainstorming sketch interface that develop a web page layout in which shapes may be left unrecognized, to a UML class diagram sketch interface, where sketchers may want to replace box-shaped classes with an index card-like image.

### B. The Framework

Rather than build a separate recognition system for each domain, it should be possible to build a single, domain-independent recognition system that can be customized for each domain. In this approach, building a sketch recognition system for a new domain requires only writing a domain description, which describes how shapes are drawn and displayed. This description is then being transformed for use in the domain independent recognition system. The inspiration for such a framework stems from work in speech recognition and compiler-compilers, which have used this approach with some success.

In this framework, the recognition system translates the domain description into a recognizer of hand drawn shapes. This is analogous to work done on compiler-compilers, in particular, visual language compiler-compilers. A visual language compiler-compiler allows a user to specify a grammar for a visual language, then compiles it into a recognizer which can indicate whether an arrangement of icons is syntactically valid. One main difference between that work and this one is that the visual language compiler-compiler deals with the arrangement of completed icons, whereas this work includes three additional levels of reasoning:

- dealing with how strokes form primitive shapes (such as lines and ellipses),
- how these primitive shapes form higher-level shapes or icons, and
- how the higher-level shapes interact to form more complicated shapes or less formal shape groups.

To build a new sketch interface:

1) A developer writes a domain description language describing information specific to each domain, including: what shapes are included in the domain, and how each shape is to be recognized and displayed (providing feedback to the user).

2) The developer will write a Java file that functions as an interface between the existing back-end knowledge system (e.g., a CAD tool) and the recognition system.

3) The User Interface customizable recognition system translates the domain description language into shape recognizers, editors, and exhibitors.

4) The UI customizable recognition system now functions as a domain-specific sketch interface that recognizes and displays the shapes in the domain, as specified in the domain description. It also connects via the Java interface (listed in Step 2) to an existing back-end system.

### C. Implementation

The framework is implemented by building:

1) a symbolic language to describe domain-specific information, including how shapes are drawn, displayed, and edited in a domain; and

2) a customizable, multi-domain recognition system that transforms a domain description language into recognizers, and exhibitors to produce a domain-specific user interface

### III. A PERCEPTUAL LANGUAGE FOR DESCRIBING DRAWING AND DISPLAY IN RECOGNITION

In order to generate a sketch interface for a particular domain, the system needs domain-specific information, indicating what shapes are in the domain and how each shape in the domain is to be recognized and displayed. Domain information should provide a high level of abstraction to reduce the effort and the amount of sketch recognition knowledge that is needed by the developer. The domain information should be accessible, understandable, intuitive, and easy for the developer to specify [7].

A shape description needs to be able to describe a generalized instance of the shape, describing all acceptable variations, so that the recognition system can be properly recognized all allowable variations. A shape description should not include stylized mannerisms (such as the number, order, direction, or speed of the strokes used) [8]that would not be presented in other sketchers' drawings of a shape, as it would require all sketchers to draw in the same stylistic manner as the developer in order for their sketches to be recognized. Thus, it has chosen to describe shapes according to their user-independent visual properties.

A perceptual language for describing shapes is being developed, for use to specify the necessary domain information. The language consists of predefined primitive shapes, constraints, and display methods. Shape descriptions primarily concern shape, but may include information such as stroke order or stroke direction, if that information would prove useful to the recognition process. The specification of editing behavior allows the system to determine when a pen gesture is intended to indicate editing rather than a stroke, and what to do in response. Display information indicates what to display after strokes are recognized.

The difficulty in creating such a language involves ensuring that the language is broad enough to support a wide range of domains, yet narrow enough to remain comprehensible and intuitive in terms of vocabulary. To achieve sufficient broadness, it was used to describe several hundred shapes in a variety of domains. Relevant figure shows a sample of the shapes described. To achieve sufficient narrowness, only perceptually-important constraints are chosen.

The language also has a number of higher-level features that simplify the task of creating a domain description. Shapes can be built hierarchically. Shapes can extend abstract shapes, which describe shared shape properties, making it unnecessary for the application designer to define these properties numerous times. As an example, several shapes may share the same editing properties. Shapes with a variable number of components, such as poly-lines or polygons (which have a variable number of lines), can be described by specifying the minimum and maximum number of components (e.g., lines) allowed. Contextual information from neighboring shapes also can be used to improve recognition by defining shape groups; for instance, contextual information can distinguish a pin joint from a circular body in mechanical engineering. Shape group information also can be used to perform chain reaction editing, such as having the movement of one shape cause the movement of another.

### IV. THE RECOGNITION SYSTEM

Recognition consists of two stages:

- stroke processing and & segmentation
- shape recognition using domain classification

During stroke processing, each stroke is broken down into a collection of primitive shapes, including line, arc, circle, ellipse, curve, point, and spiral.

During shape recognition, the properties of the strokes and shape are searched for heuristically in each domain. If a stroke or shape has multiple interpretations, all interpretations are added to the pool of recognized shapes, but a single interpretation is chosen for display. The system chooses to display the interpretation that is composed of the largest number of primitive shapes or the first found interpretation, in the case of interpretations composed of the same number of primitive shapes.

*A. Segmentation*

Segmentation refers to the process of partitioning a digital image or sketch into multiple segments (sets of pixels, also known as super pixels). The goal of segmentation is to simplify and/or change the representation of an image into something that is more meaningful and easier to analyze. Image or sketch segmentation is typically used to locate objects and boundaries (lines, curves, etc.) in images or sketches. More precisely, image segmentation is the process of assigning a label to every pixel in an image such that pixels with the same label share certain visual characteristics. The result of image segmentation is a set of segments that collectively cover the entire image, or a set of contours extracted from the image. Each of the pixels in a region is similar with respect to some characteristic or computed property, such as color, intensity, or texture. Adjacent regions are significantly different with respect to the same characteristic(s).

Several general-purpose algorithms and techniques have been developed for image segmentation. Since there is no general solution to the image segmentation problem, these techniques often have to be combined with domain knowledge in order to effectively solve an image segmentation problem for a problem domain.

The existing algorithms like the Canny Algorithm have the disadvantage of working only over the frequency values that are present in the input matrix and have a uniform threshold value. The other algorithms works on the images, that are presented as the input, that is they check for the color frequency in the image, and then segment that image on the basis of that. Not to forget as the main disadvantage of the current segmentation algorithms is the complex computations they involve. Whereas the algorithm that is presented in this document works on the direction in which the sketch is proceeding at the run time. This direction is determined by taking a pixel set (a set of 5 pixels) and processing those pixels to determine the direction. This direction determines the position or the pixel where segmentation needs to be done.

This Segmentation Algorithm is based upon the pixels or rather the pixel set (set of 5 pixels). The user draws a sketch on the console and its pixel values that is the x-coordinate and y-coordinate values (abscissa and ordinate values) are stored in the database. All the further working will be done by taking these pixel values from the database. The whole of the segmentation algorithm is divided into 8 phases. These phases are as mentioned below.

1) *Draw Sketch:* The user is given with a console on which he can draw any sketch that he requires to segment. He can draw any shape or figure with various input devices like a mouse or a light pen etc.

2) *Record Pixels:* When the user is drawing the sketch the system simultaneously records all the pixel values i.e. the abscissa and ordinate values of the drawn sketch, and stores these values into the database. When the input device is pressed or down/clicked, goes over any pixel in the consol or the canvas, highlights that pixel and the value of that pixel is stored in the database. This whole process has done at the run time, as soon as that specific pixel is highlighted and does not wait for the user to complete the sketch, thus saving processing time.

3) *Compare adjacent pixels:* After the value of the highlighted pixel is stored in the database, the system fetches these values and compares all the adjacent pixel values. This comparison is done so as to determine the flow or the direction of the sketch that is the direction in which the sketch is moving.

In order to get the direction there are 8 different categories. It means that the direction of two adjacent pixels can be one of the 8 categories. These cases are described below. Let us suppose the 2 pixels are P1 and P2, and there coordinate values are $x_1, y_1$ and $x_2, y_2$ respectively.

TABLE 1: Segmentation algorithm's 8 cases

| Case | Sketch is moving | x | y |
|---|---|---|---|
| 1 | to the positive X direction keeping Y value as constant | $x_2 - x_1 > 0$ | $y_2 - y_1 = 0$ |
| 2 | in the negative X direction keeping Y value as constant | $x_2 - x_1 < 0$ | $y_2 - y_1 = 0$ |
| 3 | in positive Y direction keeping the X value as constant | $x_2 - x_1 = 0$ | $y_2 - y_1 > 0$ |
| 4 | in negative Y direction keeping X value as constant | $x_2 - x_1 = 0$ | $y_2 - y_1 < 0$ |
| 5 | in positive X direction and negative Y direction | $x_2 - x_1 > 0$ | $y_2 - y_1 < 0$ |
| 6 | in positive X direction and positive Y direction | $x_2 - x_1 > 0$ | $y_2 - y_1 > 0$ |
| 7 | moving in negative X direction and positive Y direction | $x_2 - x_1 < 0$ | $y_2 - y_1 > 0$ |
| 8 | moving in negative X direction and negative Y direction | $x_2 - x_1 < 0$ | $y_2 - y_1 < 0$ |

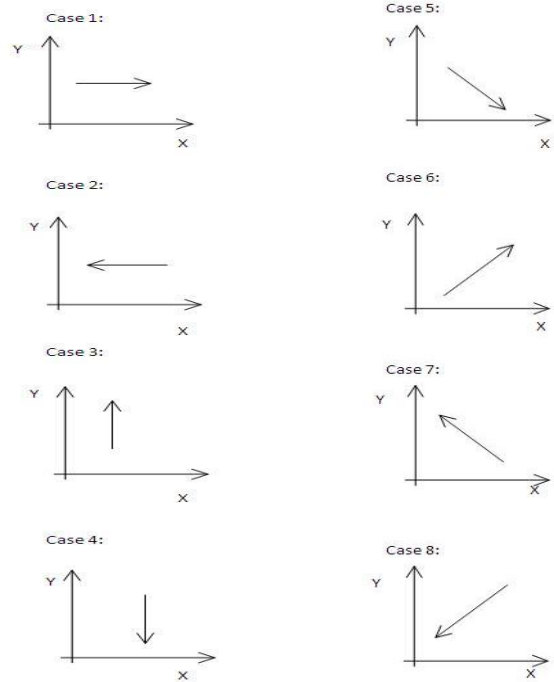

Figure 1: The various directions in which the sketch can flow and these are divided into 8 cases

B. *Domain Classification*

The Sketch is recognized using Domain Classification by the method shown here. The properties of the user drawn sketch and segments are searched heuristically in the domains and each figures of each domain. The properties of the figure in the domain and the user sketch are mapped, and finally the sketch is recognized.

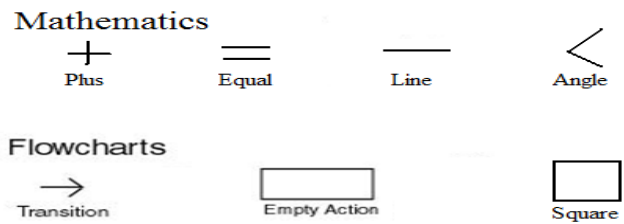

Figure 2: Various domains used by the system

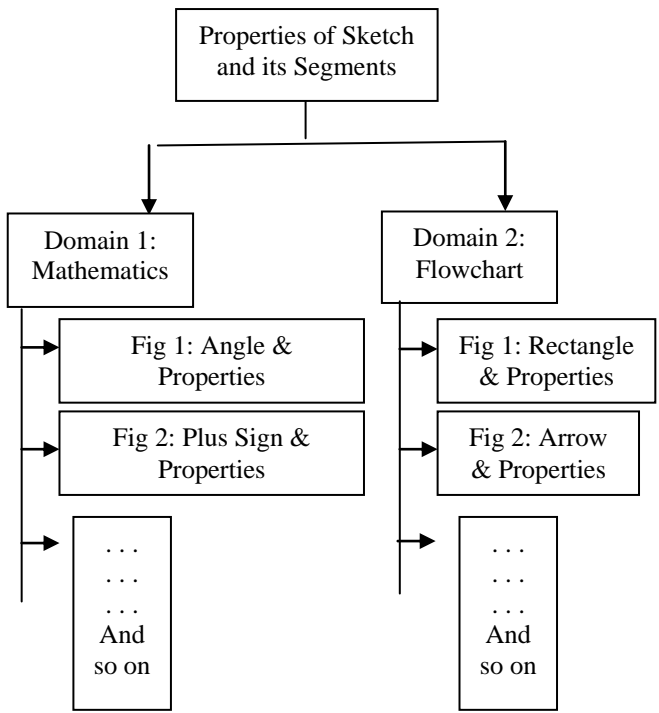

Figure 3: Domain Classification

*C. Experimental Results*

If the user draws numerous strokes rapidly, the system can slow down because there is a steady amount of time necessary to preprocess every stroke. The running time of the recognition system is analyzed and determined that, with many unrecognized shapes on the amount of time in the number of shapes on the screen to compute the property values and a logarithmic amount time to insert in to the appropriate data structure. A very small portion of time was used to do the actual recognition, even though the last portion is exponential in the number of strokes on the screen. As a result of indexing, the recognition portion takes a small amount of time, with little to no constraint calculation, as the system was only performing list comparisons. As a result, the system still reacts in what can be considered close to real-time, even with 186 shapes on the screen.

*D. Flowchart*

*1) First Module: Taking the Input*

This module has the designing and coding of the user interface. The user interface is a panel divided into frames and Java AWT elements such as buttons, textboxes, divider etc. This interface takes input drawn by mouse by the user which is a rough sketch. The input is recorded and stored in the database as pixel values with the x coordinate and y coordinate values of the highlighted pixel.

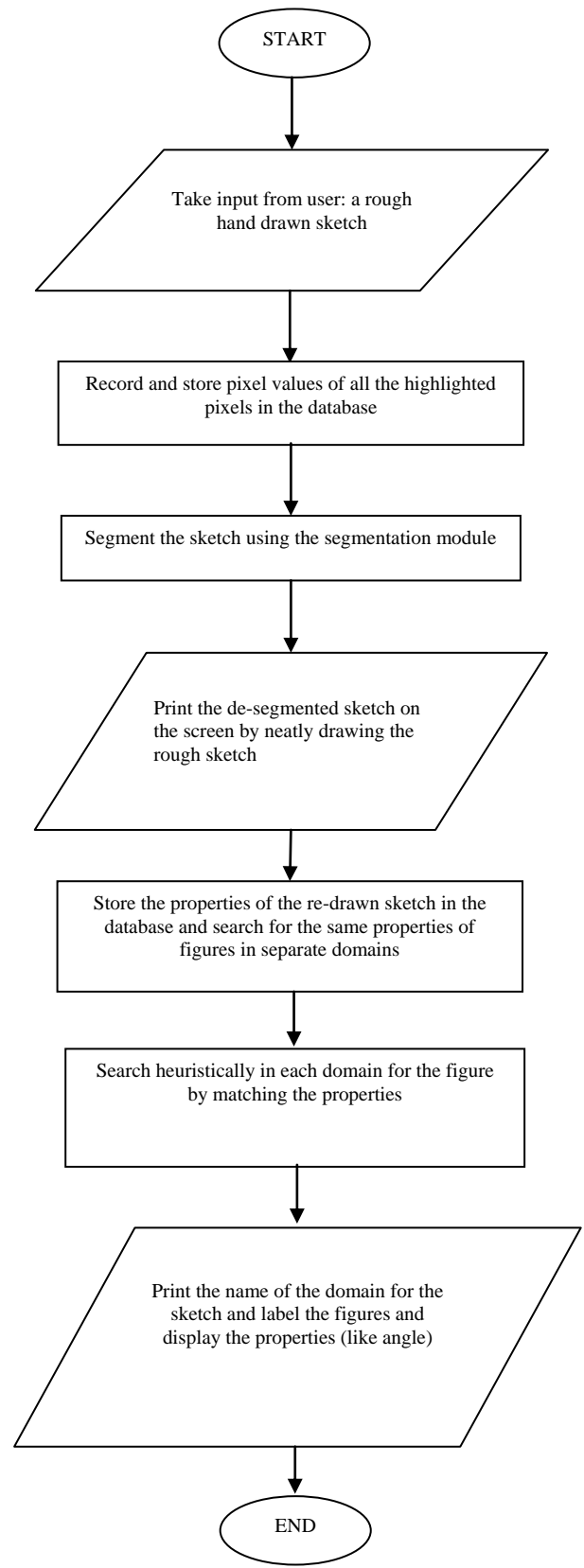

Figure 4: Flowchart for sketch recognition

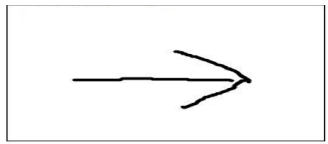

Figure 5: Hand drawn input

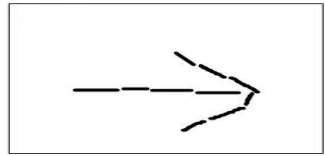

Figure 6: Processed segmented sketch

*2) Second Module: Recording the Input*

The storage of abscissa and ordinate value of the highlighted pixels starts immediately when the user clicks and drags the mouse on the interface to draw the sketch. Thus, the processing starts as user starts to draw and does not wait for the user to complete the sketch; saving the waiting time for the user.

*3) Third Module: Segmenting the Sketch*

This was one of the modules which were tough to design as the previous work done on sketch recognition does not disclose much on the segmentation algorithm.

This module also includes storing the properties of the sketch drawn and segmented so that these can be used further in recognizing the sketch.

*4) Fourth Module: Searching the sketch in the domains and the figures*

After segmentation is completed and the segmentation points are stored, the sketch has to be recognized. Domains are defined, such as mathematics, flowcharts, etc. These domains include properties of figures which lie in those respective domains.

*5) Sn*

*6) Fifth Module: Re-Drawing the Recognized figure*

After the sketch is recognized the de-segmented figure is re-drawn neatly with straight lines and proper curves. These replace the roughly drawn sketches in the user interface frame. This is done by looking at the properties of the figure such as distances and angles.

This module also includes displaying the domain name and the figure name if recognized. It may also display properties of the figure such as the angle measurements if the recognized figure is an angle. This module comes only if the sketch is recognized, if it is not recognized both domain and figure are undefined and the sketch is not re-drawn again.

*Screen Shots:*

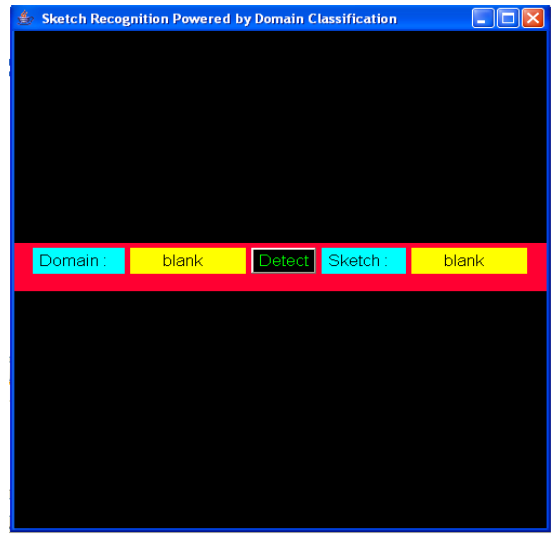

Figure 7: Basic User Interface

The Basic User Interface consists of 3 parts:
- input display
- output display
- information display

In the input display part, the user draws the sketch. In the output display the recognized sketch is displayed. Whereas in the information display shows the information like in which domain the sketch lies and what is the name of that sketch.

Following figure represents a sketch that is drawn by the user. When the user press the detect button, the system does its processing and generates an output.

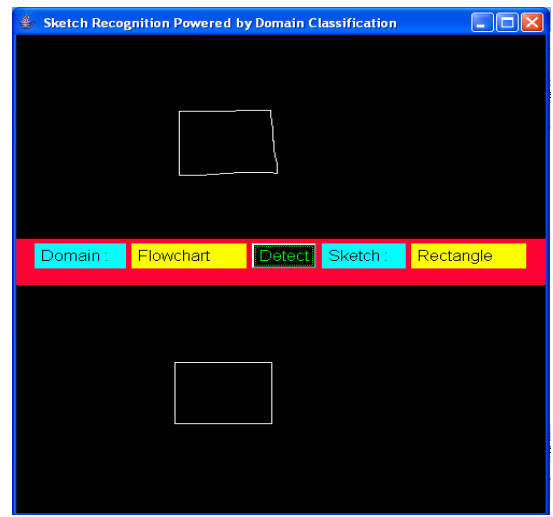

Figure 8: User Interface with input sketch and recognized output sketch as a Rectangle

This figure represents an input as submitted by the user in the previous figure. Along with that there is the recognized figure with proper orientation. Also there is the domain i.e. "Flowchart" in which this sketch lies and the name of this sketch is "Rectangle".

recognized because the sketch which the user has drawn doesn't present in any of the domains. Thus the system returns a blank output space and domain and sketch value as "Undefined".

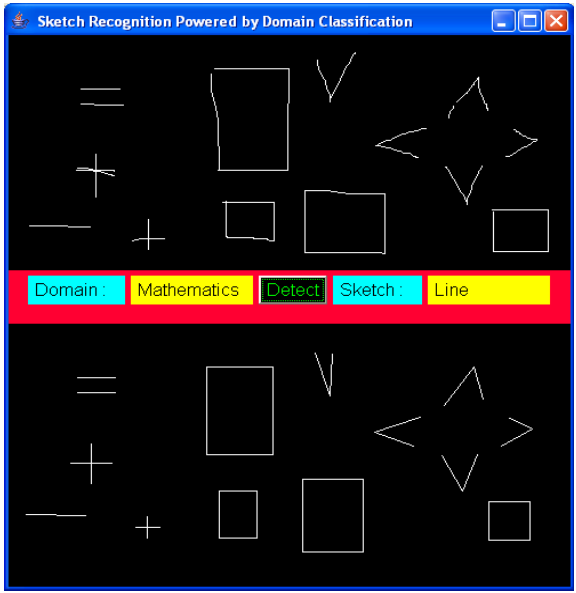

Figure 9: User Interface with input sketchs and recognized output sketchs in a single window

Figure 9 is a special figure. In this figure, it is represented that the user can draw any number of sketches in the input space and when he presses the Detect button, all of the input sketches are detected. Also for each input sketch a correct recognized sketch is displayed in the output space.

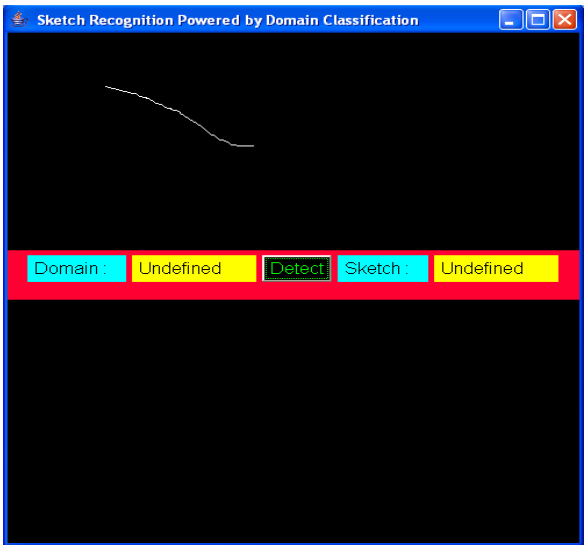

Figure 10: User Interface with input sketch only because this sketch is not recognized by the system.

The figure 10 represents a sketch that is drawn by the user. When the user press the detect button, the system does its processing and generates an output. In this figure no sketch is

Figure11: it shows the database table Sketch1

The figure 11 shows the values that are contained by the dynamically created table Sketch1. This table consists of the field ID, X and Y co-ordinates.

Figure 12: it shows the database table Sketch1CAT

The figure 12 shows the values that are contained by the dynamically created table Sketch1CAT. This table contains the fields ID and the Cat which stores the category value.

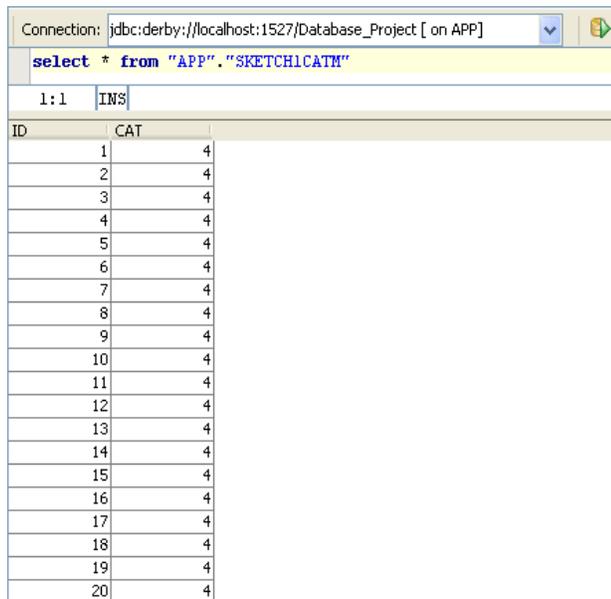

Figure 13: it shows the database table Sketch1CATM

Figure 13 shows the values that are contained by the dynamically created table Sketch1CATM. This table overwrites the field of Category in a pixelset.

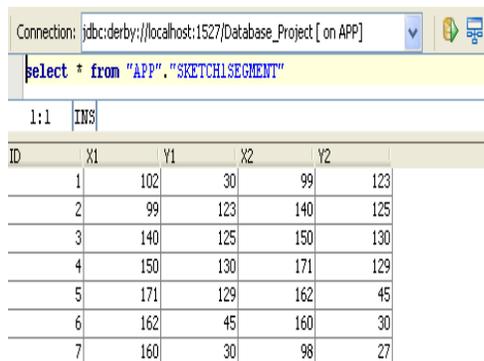

Figure 14: it shows the database table Sketch1CATM

Figure 14 shows the values that are contained by the dynamically created table Sketch1Segment. This table consists of all the values of the segments i.e. the segment starting and ending point along with the segment number.

V.  MERITS AND DEMERITS

*A. Merits*
- This system can be used for multiple sketches at the same time.
- Segmentation Approach used removes the need of use any filtration algorithm, making it faster.
- Has an easy to use user interface.
- This interface can be connected to other applications which need recognition modules.
- It is platform independent.

*B. Demerits*
- Currently the system cannot be used for figures with curves, however there is an algorithm designed by us which can be implemented further.

VI.  FUTURE WORK

It can be used to develop:

- Handwriting Recognition Systems.
- Fast Flowchart Designers.
- Other Architecture Designers.
- Higher accessibility tools for users to use the Interface more efficiently. (for OS etc.)
- Same approach can be used to implement similar projects for touch screen applications on various new upcoming OS(s) like nokia rim,apply iphone,windows mobile android,google garnet,palm web bada os , maemo os, meego os etc.

VII CONCLUSION

The over-arching goal of this work is to make human-computer interaction as natural as human-human interaction. Part of this vision is to have computers understand a variety of forms of interaction that are commonly used between people, such as sketching. Computers should, for instance, be able to recognize the information encoded in diagrams drawn by humans, including mechanical engineering diagrams.

Ordinary systems offer one the freedom to sketch naturally, but it does not provide the benefits of a computer-interpreted diagram, such as more powerful design advice or simulation abilities. Sketch recognition systems bridge that gap by allowing users to hand-sketch their diagrams, while recognizing and interpreting these diagrams to provide the power of a computer-understood diagram. Many sketch systems have been built for a particular domain. Unfortunately, these sketch systems may not fill the needs of the sketcher, and building these sketch systems requires not only a great deal of time and effort, but also an expertise in sketch recognition at a signal level [7]. Thus, the barrier to building a sketch system is high. This researcher wants to empower user interface developers, including designers and educators, who are not experts in sketch recognition, to be able to build sketch recognition user interfaces for use in designing, brainstorming, and teaching.

In response to this need, the framework is developed for facilitating User Interface development. As part of the framework, this project has developed a perception-based sketching language for describing shapes, and a customizable recognition system that automatically generates a sketch recognition system from these shapes. In order to allow drawing freedom, shapes are recognized by what they look like, rather than by how they are drawn. This language provides the ability to describe how shapes in a domain are drawn and displayed within the user interface.

Because humans are naturally skilled at recognizing shapes, the system uses human perceptual rules as a guide for the constraints in the language and for recognition.

The sketch recognition system's accuracy can be improved by
- Combining user-dependent (feature-based) recognition with user-independent (geometric) recognition techniques
- Incorporating global and local context into the recognition system, including geometric, perceptual, functional, multi-modal, similarity, and common sense context.

AUTHORS' PROFILE

**Vasudha Vashisht** received her bachelor's and master's degree in Computer Science from M.D. University, Haryana, India. She has 6 years of experience in teaching. Currently, she is working as a Assistant Professor in the Dept. of Computer Sc. & Engg. at Lingaya's University, Faridabad, Haryana, India. She has authored 10 papers and her areas of interests include artificial intelligence, Cognitive Science, Brain Computer Interface, Image & Signal Processing. Currently she is pursuing her doctoral degree in Computer Science & Engg. She is a member of reputed bodies like IEEE, International Association of Engineers, International Neural Network Society, etc.

**Tanupriya Choudhury** received his bachelor's degree in CSE from West Bengal University of Technology, Kolkata, India, and master's Degree in CSE from Dr. M.G.R University, Chennai, India. He has one year experience in teaching. Currently he is working as a Senior Lecturer in dept. of CSE at Lingaya's University, Faridabad, India. His areas of interests include Cloud Computing, Network Security, Data mining and Warehousing, Image processing etc.

**Dr. T. V. Prasad** received his master's degree in Computer Science from Nagarjuna University, AP India and a doctoral degree from Jamia Milia Islamia University, New Delhi, India. With over 16 years of academic and Professional experience, he has a deep interest in planning and executing major IT projects, with deep interest in research in CS/IT and bioinformatics. He is the author of 60+ journal/conference/book chapter/white paper publications.He has also held respectable positions such as Deputy Director with Bureau of Indian Standards, New Delhi.His areas of interest include bioinformatics, artificial intelligence, consciousness studies, computer organization and architecture. He is a member of reputed bodies like ISRS, CSI, APBioNet, etc.